# Knowledge Engineering in the Long Game of Artificial Intelligence: The Case of Speech Acts

**Marjorie McShane**                    MARGEMC34@GMAIL.COM
**Jesse English**                      DRJESSEENGLISH@GMAIL.COM
**Sergei Nirenburg**                    ZAVEDOMO@GMAIL.COM
Cognitive Science Department, Rensselaer Polytechnic Institute, Troy, NY, 12180, USA

## Abstract

This paper describes principles and practices of knowledge engineering that enable the development of holistic language-endowed intelligent agents that can function across domains and applications, as well as expand their ontological and lexical knowledge through lifelong learning. For illustration, we focus on *dialog act modeling*, a task that has been widely pursued in linguistics, cognitive modeling, and statistical natural language processing. We describe an integrative approach grounded in the OntoAgent knowledge-centric cognitive architecture and highlight the limitations of past approaches that isolate dialog from other agent functionalities.

## 1. Introduction

It is a truism that general artificial intelligence (AI) cannot be achieved in one fell swoop. So, studies about intelligence are inevitably chopped into project-sized bites. The chopping is not easy, but it is made easi*er* by the field's acceptance of certain chopping practices. For example, some developers work on a **single module** (architecture, phenomenon, subsystem) of a comprehensive AI system, often assuming the availability of unattainable prerequisites and envisioning future overall system integration of an unspecified nature. Other developers work in a **single small domain**, addressing multiple modules and their interactions but not the various preconditions for transcending their domain of choice. Still others work on a **single application,** focusing on its utility without concern for its contribution, if any, to long-term progress in developing comprehensive AI systems. There are both scientific and practical justifications for organizing research and development in these ways. However, if we are to make progress in the long game of achieving human-level AI, these simplifying methodologies cannot be pursued forever. The reason why is that the solutions to simplified problems are not additive. Imagining that they *are* would be like saying that skyscraper technology equals tent technology plus log cabin technology plus two-story suburban home technology.

In the OntoAgent program of research we are working toward developing artificial intelligence in domain-independent, human-like, language-endowed intelligent agents (LEIAs) (McShane & Nirenburg, 2021). A core requirement is **holistic knowledge engineering that takes a long view** – i.e., that foresees a wide range of interconnected agent capabilities across domains and applications. In this paper, we motivate the need for this take on knowledge engineering by juxtaposing its benefits with the limitations of more insular approaches pursued in the past. To focus the discussion, we concentrate on an issue that has been addressed from quite disparate angles: *dialog modeling* –also called *dialog act modeling*, and *speech act modeling* (we will use the terms interchangeably).

Broadly speaking, dialog modeling involves using expectations about how dialogs typically proceed – e.g., questions are usually followed by answers – in order to improve some aspect of language processing or other agent functioning. It has been invoked in programs of research and development that reflect all three kinds of simplifications listed above. In descriptive and theoretical linguistics – later adopted into computational linguistics – dialog modeling has been studied in isolation, separated both from the semantic content subsumed within the speech acts and from non-speech acts that freely cooccur with the language. In cognitive systems development, speech acts are commonly paired with specific semantic propositions, resulting in domain-specific language modules that lack generalization and cross-domain portability. And in machine-learning-based AI, various inventories of speech-act tags have been used to inform supervised machine learning; this gives applications the veneer of linguistic grounding but with no actual linguistic sophistication or movement toward language understanding or generalized artificial intelligence. Importantly, none of these approaches to speech acts offers a path toward broad-coverage, explainable, human-like cognitive systems.

The main claims of the paper are as follows: (a) computational cognitive modeling needs to be approached with a long, integrationist view, and (b) this long view fundamentally changes how we conceive of phenomena and solve problems. The paper will unfold as follows. First we will explain the linguistic origins of isolated dialog act modeling and how this approach was adopted rather too literally into computational linguistics – to the ultimate disservice of the latter. Then we describe one way in which dialog act tags have been used in statistical AI, leading to improvements in the performance of individual systems but without advancing natural language processing overall. Finally, using the example of dialog treatment as a springboard, we provide a glimpse into how knowledge engineering in OntoAgent transcends the limitations of current approaches.

## 2. The Precedent for Isolating Speech Acts and Dialog Models

Divide and conquer has a venerable place in the history of science. Since at least the times of Descartes, the scientific method has become more or less synonymous with the analytical approach, whereby a phenomenon or process is decomposed into contributing facets or components. The idea is that each component is first studied independently, followed by a synthesis step that gives rise to a comprehensive explanation of the phenomenon or process. Unfortunately, practice shows a tendency toward permanent isolation, which artificially constrains the purview of theories and the scope of models, and leads to the indefinite postponement of the all-important synthesis step.

In linguistics, dialog acts have long been studied as an aspect of pragmatics (see the literature reviews in, e.g., Traum, 1999a, 2000). As Traum (2000: 7) says, "When engaging in a study related to dialogue pragmatics, a researcher is confronted with a bewildering range of theories and taxonomies of dialogue acts to choose from." As is common in theoretical and descriptive linguistics, dialog acts have been neatly shaved off from two intimately connected phenomena: the meaning of the propositions scoped over by those dialog acts, and any actions outside of language that are relevant to the situation, such as responding to an utterance by shrugging. So, dialog act models address the fact that when someone asks a question, the interlocutor typically answers it, but they say nothing about what is actually asked. As Traum (1999b, p. 1) writes, "In studying speech acts, the focus is on pragmatics rather than semantics – that is, how language is used by agents, not what the messages themselves mean…"

Between around 1980 and the early aughts, computational dialog modeling was studied in earnest, primarily as an aspect of planning. For example, Allen & Perrault (1980) put forth a goal- and plan-based approach to dialog processing, influenced by classical AI approaches to planning. Later work in dialog processing (e.g., Lemon & Gruenstein, 2004) shifted to relying predominantly on dialog cues – still isolated from semantic content. In addition to theoretical work, that period also gave rise to tools-related work: e.g., DIPPER (Bos et al., 2003) claimed to combine the strengths of the goal-oriented and cue-oriented paradigms. It used "aspects of dialogue state as well as the potential to include detailed semantic representations and notions of obligation, commitment, beliefs and plans" (Bos et al., 2003). Key here is that the DIPPER infrastructure enabled various things, such as semantic representations, but did not address their computation or how their computation might actually inform decisions about the infrastructure itself.

The drawbacks of artificially isolating linguistic modules can be illustrated using the model of grounding (a class of dialog acts) reported in Traum (1999a). Traum's model of grounding involves a state transition table that indicates, e.g., that if the dialog state requires an acknowledgement by the listener (an indication that he or she understands or is listening), but the speaker keeps on talking without that acknowledgment happening, then that dialog is ungrounded. Although Traum's model is formal and nicely captures human intuitions, it has a lot of outstanding prerequisites – as he himself points out. For example, the model does not address how to divide up chunks of speech that should be subject to grounding; it does not offer a method of automatically identifying which grounding act was performed; it does not attempt to account for what utterances mean, which is the only way that one can determine whether a grounding act is occurring; it does not account for speaker intention, which is an aspect of mindreading that is needed to judge whether grounding is needed at a given moment and, if so, what kind; and it assumes a binary distinction between grounded and ungrounded, which is too coarse-grained. In short, the model is intuitively satisfying for what it covers but it doesn't address a wide enough scope of phenomena to be implementable in agent systems.

The past two decades have witnessed ever more work on integrated cognitive systems that emulate human performance, which makes the separation of dialog processing from other agent capabilities increasingly less justified. For example, McShane and Nirenburg (2009) explain why a dedicated dialog model is not needed in the OntoAgent application called Maryland Virtual Patient, a proof-of-concept clinical tutoring system (McShane & Nirenburg, 2021, Chapter 8). In this paper, we stick to that claim and extend the modeling and associated argumentation in two ways. First, we detail how knowledge and system engineering can be made largely domain-independent in order to enable agents to be easily ported across domains (the 2009 paper involved just one application). Second, we show how anticipatory behavior can be operationalized, which not only renders intelligent agents more human-like, but also offers complexity reduction for things like lexical disambiguation and fragment interpretation.

To close out this historical perspective, we must note that it is not by chance that our examples of dialog modeling by others are from two decades ago. Around that time, the statistical turn in NLP was so sharp that practically all interest and resources turned away from knowledge-based systems to knowledge-lean ones. Despite some later experiments based on the work cited above (e.g., Allen et al., 2007), to our knowledge there have been no theoretical or descriptive advancements *that can inform our work*.[1] One recent trend has been to incorporate dialog act tags into knowledge-lean systems, a topic to which we now turn.

## 3. Dialog Act Tags in Knowledge-Lean Systems

Over the past few decades, the vast majority of work on AI, including NLP, has involved machine learning (ML), to the point where AI has become *de facto* synonymous with ML. However, ML is unlikely to solve the core issue of language *understanding* for reasons detailed in McShane and Nirenburg (2021). Although the limitations of ML as applied to NLP – what we will call ML-NLP – have long been clear to anyone with even a passing knowledge of linguistics, NLP practitioners have found it expedient to focus on near-term results and postpone tasks and applications that inevitably rely on the computation of meaning. This is not surprising considering that fundamentally treating natural language requires tackling the complexity of human cognition. We devote a section of this paper to ML-NLP not because it informs our main story (it does not) but for two other reasons. First, since ML-NLP is the mainstream approach, everything *other* is expected to be compared with it. Second, until knowledge-based NLP regains its proper place in the field of AI, the need for knowledge-based approaches must be tirelessly argued for.

Of the various genres of ML, supervised learning is the most relevant to our discussion because it uses human-interpretable features. Practitioners of supervised ML-NLP have long argued that the core prerequisite for improving their currently modest results is enhancing the inventory of features. They call this judicious selection of distinguishing features on which to base comparisons and classifications. As concerns dialog acts in particular, they are so widely addressed in ML-NLP that already a decade ago this was the subject of a detailed survey article (Král & Cerisara, 2010).

In broad strokes, given a particular ML engine, the more features it uses and the larger the annotated corpus it invokes, the better the results should be. As Manning (2004) puts it, "In the context of language, doing 'feature engineering' is otherwise known as doing linguistics. A distinctive aspect of language processing problems is that the space of interesting and useful features that one could extract is usually effectively unbounded. All one needs is enough linguistic insight and time to build those features (and enough data to estimate them effectively)." To paraphrase, Manning's claim is that progress on NLP is limitless if one piles in more features and has people annotate ever larger corpora for them. We disagree on the following counts.[2]

---

[1] There is plenty of work on dialog and NLP that contributes to the field but neither informs our work nor has a place in the story we are telling – which is necessarily highly selective in this short space. As regards the pitfalls and dangers of machine learning – particularly, deep learning – interested readers can find many published analyses: e.g., Babic et al. (2021), Hutson (2021), and Marcus (2018).

*3.1. No matter the feature inventory, ML – as currently approached – manipulates words, not their meanings.* Words are to meanings like the tip of the iceberg is to the vast expanse of ice below the surface. The ML community has used words as a proxy for meanings under one of two assumptions: (a) [the naïve assumption] words are sufficient proxies for their meanings, or (b) [the linguistically informed but defeatist assumption] words are *not* sufficient proxies for their meanings but they are the best we've got – at least given the widespread commitment to ML as the method of choice. But no matter how many "surfacy" linguistic features are pumped into ML-oriented NLP systems, the latter will not achieve human-like language abilities as long as the features are attaching exclusively to words, not the contextual meanings of propositions. Historically, the most widely used and commonly annotated features for ML have been syntactic, which renders the word vs. meaning distinction not central. However, when annotation practices expanded into the realms of semantics and discourse, the import of the word vs. meaning distinction should have been addressed more fundamentally. Alas, it wasn't, leading to the linguistically bizarre situation where a particular semantic or discourse feature is inserted into an otherwise unanalyzed sequence of words. All such tagging – including dialog act tagging – gives systems no more than a veneer of linguistic grounding.

*3.2. Only simple linguistic features and values are ever annotated in support of supervised ML.* Given real-world constraints of time and cost, people can only reliably annotate the simplest of linguistic phenomena. (One cannot disregard the time and cost burden of annotation because the time and cost of knowledge acquisition have been at the heart of the ML community's arguments against knowledge-based systems!) Annotation guidelines concentrate on what is easy to annotate and do not touch difficult phenomena. The relative successes of supervised ML for the easy cases (essentially, syntax and "light" semantics/pragmatics) cannot be recreated for the more difficult cases (see McShane & Nirenburg, 2021, Section 1.6.12 for discussion). Moreover, even if large corpora could, somehow, be annotated for more difficult phenomena, it is not at all clear that this would lead to any useful ML. For a glimpse why, consider the example of bridging references, in which a definite description (a noun phrase with 'the') can be used without a coreferential antecedent if it is virtually introduced into the context by a different concept. For example, in the sentence "Our local pool was closed because *the filter* was broken" *the filter* is in a bridging relation with *pool*; we are not surprised by the mention of '*the* filter' because it was pulled into the conceptual space triggered by the pool. This *pool ~ filter* correlation may or may not occur a single time in a corpus, and it will give an ML system no insight into resolving the parallel usage of *cake ~ icing* in "I had a piece of cake for lunch but didn't like the icing". To understand what is going on in these examples, a system needs to (a) know that bridging references explain some uses of 'the', (b) know that they rely on the ontological HAS-AS-PART relation, and (c) have the ontological knowledge that pools have filters and cakes have icing. In short, supervised ML has severe practical limitations that never seem to see the light of day in the literature.

*3.3. Improvement of an application system does not imply advancement of the state of the art overall.* ML is used in AI because it allows for the quick development of certain kinds of applications that people find useful. ML-oriented development is all about making applications work better. If using a particular feature improves the system, it's in; if it makes the system worse, it's out. Its presence or absence need not make sense to people. This practical view of features is entirely appropriate for application-oriented work. However, it does not imply scientific advancement toward human-level intelligence.

A nice illustration of application-driven dialog-act selection is found in Stolcke et al. (2000).[3] Their inventory of forty-two dialog acts was seeded by the Dialogue Act Markup in Several Layers (DAMSL) tag set (Core & Allen, 1997) but they modified it to suit the specifics of their corpus – the dialogs in the Switchboard corpus of human-human conversational telephone speech (Godfrey et al., 1992). If Stolcke's inventory of dialog acts is viewed in isolation, some aspects of it are rather perplexing. For example, the inventory contains only a small and idiosyncratic number of performative speech acts – APPRECIATE, APOLOGY and THANKING – and it includes some entities that are arguably not speech acts at all, such as OR-CLAUSE and QUOTATION. However, when viewed in the context of their chosen application – improving conversational speech recognition by whatever means – the inventory is less surprising. In fact, Stolcke et al. underscore their practical aims, saying that they "decided to label categories that seemed both inherently interesting linguistically and that could be identified reliably. Also, the focus on conversational speech recognition led to a certain bias toward categories that were lexically or syntactically distinct (recognition accuracy is traditionally measured including all lexical elements in an utterance)" (p. 343). In short, their goal was to achieve correct speech recognition: whatever features helped that, for whatever reason, were useful contributors. Similarly application-specific dialog act inventories are found in other systems. For example, the Verbmobil-1 project used dozens of speech-act tags, including BYE, FEEDBACK, DIGRESS, and MOTIVATE, none of which are found in the 12-tag inventory used for the Map Task (both cited in Jurafsky, 2006). Even more domain-specific are the dialog acts in Jeong and Lee's (2006) flight reservation application, which considers "Show Flight" – a combination of a request and its semantic content – to be a dialog act.

No doubt, particular applications can benefit from bespoke feature sets, including those associated with dialog acts, but the use of such features to improve particular application systems does not contribute to solving the core problems of dialog processing in language understanding systems.

*3.4. Coarse-grained linguistic labels obscure fine-grained linguistic realities.* Consider the example of WH-QUESTION, a dialog act that is included in many dialog-act inventories. As we discussed in Section 3.1 above, recognizing the dialog act WH-QUESTION does not involve determining the meaning of that question. Nor does the recognition process address the fact that WH-QUESTIONs can be encoded using syntactic structures of varying complexity, from quite simple ones to those including long-distance dependencies, e.g., *Who came? Who did you say had to come? Who did you say was the first to be invited to come?* Consider also that these same kinds of questions can be asked using formulations that cannot be trivially recognized by a sentence-initial wh-word. For example, the following are indirect speech-act formulations of the questions above: *I need to know who came. You said somebody had to come; tell me who. It would be great if you'd tell me who the first person you invited to come was.* The point is that tags like WH-QUESTION are as linguistically impoverished as they are easy to automatically recognize.

---

[3] This example is discussed in more detail in McShane and Nirenburg (2021: 47-48).

Let us reiterate the point of this section. Sending an ever-larger inventory of features to a ML engine is not doing linguistics and does not represent progress toward human-level natural language processing. Returning to the skyscraper metaphor, you don't get a skyscraper by piling up ever more steel, concrete and glass – careful construction is needed. We now turn to such construction in our work on developing OntoAgents.

## 4. A Snapshot of OntoAgent

The OntoAgent architecture (Figure 1) is comprised of the following components:

- Two input-oriented components: **perception and interpretation**. No matter how the agent perceives something – be it through language, vision, interoception, etc. – it must interpret the incoming data in terms of its knowledge bases, resulting in ontologically-grounded *knowledge* that is stored to memory.
- The internal component covering **attention and reasoning** which, among many other things, is responsible for learning and planning future actions.
- Two output-oriented components: **action specification and rendering**. Action specification involves deciding what to do, whereas rendering makes it happen. For example, in response to a yes-no question, the LEIA might decide to respond positively and to convey that response using an utterance, such as "Yes", "Uh-huh", or "I sure did".
- A supporting service component: **memory and knowledge management**.

We call OntoAgent a *knowledge-centric* architecture because, as a first priority, we are addressing the need for large-scale knowledge – in contrast to uninterpreted data, of which there is a surplus (Nirenburg, McShane & English, 2020).

All knowledge bases and meaning representations are recorded using an ontologically-grounded metalanguage. The ontology is a hierarchical, inheritance-supported graph of concepts – OBJECTs and EVENTs – each of which is described using an average of 15 PROPERTies.[4] Ontological concepts are not English words in upper-case semantics: they are unambiguous entities, written in a formal metalanguage, whose meaning is understood as the meaning of all of the property-facet-value triples describing them.

---

[4] See Nirenburg and Raskin (2004) for motivation for this style of ontology for intelligent systems. See McShane & Nirenburg (2021) for a discussion of why a metalanguage is needed – i.e., why a natural language is not good enough.

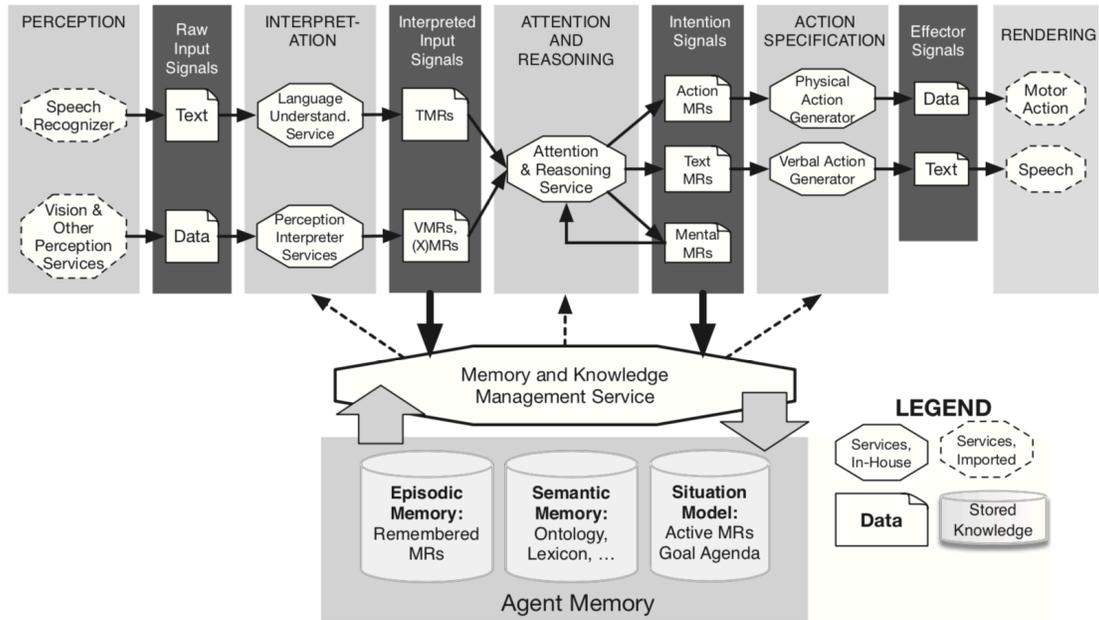

Figure 1. The OntoAgent cognitive architecture.

The ontology's *property-facet-filler* descriptions can be illustrated by the short excerpt from the frame for SURGERY in Figure 2.

| SURGERY | | |
|---|---|---|
| IS-A | value | MEDICAL-EVENT |
| AGENT | default | SURGEON |
| | sem | PHYSICIAN |
| | relaxable-to | HUMAN |
| BENEFICIARY | default | MEDICAL-PATIENT |
| | sem | ANIMAL |
| INSTRUMENT | sem | SCALPEL, SCISSORS, FORCEPS, … |
| CAUSED-BY | sem | MEDICAL-DISEASE, INJURY |
| *etc.* | | |

Figure 2. An excerpt from the frame for SURGERY in the LEIA's ontology

This grain-size of knowledge supports many kinds of agent reasoning about language and the world such as:

- The disambiguation of argument-taking words during language processing: In *The surgeon just started the operation,* 'operation' likely means SURGERY, not MILITARY-OPERATION

- Reasoning about reference and coreference: e.g., bridging references can be explained by meronymy: *When I went into her room the window was opened*
- Reasoning during new-word learning: e.g., when a doctor uses the unknown word 'achalasia' in *I think you have achalasia,* the LEIA can guess that *achalasia* is a kind of DISEASE.

The LEIA's ontology currently contains knowledge like this for around 9,000 concepts, mostly from the general domain (sometimes referred to as an *upper ontology*).

The ontology also contains far more detailed knowledge about particular domains, recorded as complex events, or *scripts*. For example, for the Maryland Virtual Patient system (McShane & Nirenburg, 2021, Chapter 8), we developed scripts to support interactive physiological simulations of virtual patients, their conversations with the physician about their care, and the functioning of a virtual tutor. Developing scripts requires a far greater sophistication of knowledge engineering than is needed for the basic ontology, as well as the integration of static knowledge with knowledge about how to apply it.

At a descriptive grain-size between basic concepts and scripts we have scriptlets. Like scripts, scriptlets include coreferenced variables and, as needed, calls to procedural routines. But like the broad-coverage, ontological knowledge, they can be acquired fast and they reflect relatively free-standing knowledge chunks. Scriptlets will be discussed further in the next section.

The LEIA's current ontology contains the kinds of knowledge it needs not only to operate in application systems, but also to bootstrap the learning of more ontological knowledge. This style of ontology is, in our experience, more useful to LEIAs than other options that have been pursued in the past because it responds to the needs of the agent systems that use it.[5]

To mediate between language and ontology, LEIAs use a computational lexicon that contains interlinked syntactic and semantic descriptions of words, phrases, and constructions. The semantic descriptions are written in the metalanguage of the ontology. Consider, for example, the construction *someone performs surgery on someone*, which we present here without the formalism used in our lexicon (we introduce the formalism in the next section*).* The lexical sense describing this construction includes a syntactic zone that says that the expected constituents are 'NP-Subject *perform surgery on* NP-Object".[6] The semantic zone says that this instantiates the concept SURGERY, whose AGENT is the meaning of NP-Subject, and whose THEME is the meaning of NP-Object. So, if the LEIA receives the input *Yesterday Dr. Jones performed surgery on Mary Smith,* it will generate the following text meaning representation, whose components are numbered instances of ontological concepts:

---

[5] For example, although we applaud the CYC team's efforts to build a realistic-size ontology, its organization of knowledge as a "sea of assertions" complicates its use in machine reasoning (see McShane & Nirenburg, 2021, Section 1.4.4.1, for discussion and related references).

[6] NP refers to noun phrase.

SURGERY-1
    AGENT                    HUMAN-1
    THEME                    HUMAN-2
    TIME                     ((find-anchor-time) -1 DAY)[7]
HUMAN-1
    HAS-TITLE               Dr.
    HAS-SURNAME         Jones
HUMAN-2
    HAS-PERSONAL-NAME   Mary
    HAS-SURNAME         Smith

Figure 3. The text meaning representation for *Yesterday Dr. Jones performed surgery on Mary Smith.*

Generating ontologically-grounded structures like this is what we mean by *interpreting* language inputs. How are they better than sentences of language? They reduce paraphrases to a single representation, they reflect the system's decisions about ambiguity resolution and the reconstruction of ellipsis, they present the results of reasoning about implicatures, and much more. LEIAs interpret all inputs prior to saving the interpretations to memory. They then rely on their remembered knowledge for reasoning, learning, and decision-making.

Given a very simple example like the one above, language interpretation might seem easy. But, in fact, it is so hard that mainstream AI gave up on it some 30 years ago. However, we believe it must remain on agenda because AI systems will never approach human-like capabilities unless we solve the problem that the words in a text are only the tip of the iceberg when it comes to their contextual meaning. Readers interested in a more comprehensive overview of OntoAgent, with an emphasis on language understanding, are encouraged to peruse *Linguistics for the Age of AI* (McShane & Nirenburg, 2021), which is available open access at https://direct.mit.edu/books/book/5042/Linguistics-for-the-Age-of-AI.

## 5. Dialog Folded Into Overall OntoAgent Modeling

At this point in our story we have established several things. For comprehensive agent modeling, it is inadvisable to continue to (a) isolate dialog acts from their meanings, (b) create models with a large inventory of unfulfillable prerequisites, (c) separate dialog acts from other actions, including non-verbal ones, or (d) hope that ML-style dialog act detection will support progress toward human-level language processing. Add to these an axiom supported at length in McShane and Nirenburg (2021): small-domain cognitive systems will not grow up to broad-coverage ones because, at least as concerns language, removing domain- and application-oriented simplifications opens the floodgates of complexity.

In this section, we briefly present seven aspects of knowledge engineering that are salient for dialog processing and general intelligent agent development. There is, unfortunately, no perfect ordering of these issues so we use forward and backward cross-references as needed.

---

[7] This is a call to a procedural semantic routine which, if run, will attempt to find the day when the statement was made and then calculate back one day prior, thus grounding the meaning *yesterday*.

*4.1. All events are treated similarly: speech acts are not special and there is no need for a dedicated dialog model.* The reason is that speech acts liberally interact with non-speech acts, as is clear from everyday examples: If X asks Y a question (dialog act) then Y might answer (dialog act) or shrug (non-dialog act). If X tells Y a joke (dialog act) then Y might laugh (non-dialog act) or say it isn't funny (dialog act). And if X punches Y (non-dialog act) then Y might yell at him (dialog act) or punch him back (non-dialog act).

*4.2. To the degree possible, chunks of ontological knowledge are recorded as scriptlets.* We define scriptlets as compact ontological descriptions for which coreferencing of case-roles is required. (By contrast, "basic" ontological knowledge involves property-facet-filler triplets without coreferenced variables, and full-fledged domain scripts can be of any size and complexity.) Often scriptlets involve a single pair of moves with variations, as in the example below, which is the concept for asking a *yes-no* question.[8]

| REQUEST-INFO-YN | | |
|---|---|---|
| AGENT | value | HUMAN-#1 |
| BENEFICIARY | value | HUMAN-#2 |
| HAPPENS-NEXT | default | RESPOND-TO-REQUEST-INFO-YN-#1 (AGENT HUMAN-#2) (BENEFICIARY HUMAN-#1) |
| | sem | EVENT-#1 (AGENT HUMAN-#2) |

The first thing to say about this structure is that, even though asking a *yes-no* question might seem like a linguistic (not ontological) phenomenon, this is actually an ontological concept for the following reasons: (a) one can ask a *yes-no* question not only linguistically but by using gestures – e.g., I can ask you if you want a muffin by pointing to it and raising my eyebrows; (b) cross-linguistically, there are many linguistic expressions as well as many gestures that can express a *yes-no* question; and (c) this kind of question sets up expectations about what will follow (HAPPENS-NEXT), and those expectations are a proper part of ontology.

The key information in this scriptlet is what is likely to happen next after one person, HUMAN-#1, asks another person, HUMAN-#2, a *yes-no* question. The default value of HAPPENS-NEXT is for HUMAN-#2 to respond in a typical way for *yes-no* questions. That inventory of responses, which can be verbal or non-verbal, will be recognized as instances of the concept RESPOND-TO-REQUEST-INFO-YN in ways described below. However, apart from answering the question, the interlocutor can also carry out some other action, like asking a clarification question or laughing – options that are covered by the EVENT option in the *sem* facet.

There are three reasons why we consider scriptlets a core aspect of knowledge engineering for cross-domain, human-like intelligent agents:

1. Scriptlets record knowledge in a way that we believe is psychologically plausible – i.e., as small information packages that should not be bound to weighty domain scripts. If a group of random people was asked, out of the blue, "What typically happens next when somebody picks up a hammer / sits in the driver's seat?" the answers would very likely be "He or she hits a nail / turns on the ignition." People and systems don't need to know all of the details about building things or driving a car to have and use this level of knowledge.

---

[8] Lexical constructions for tag questions also map to this concept since they expect the same kinds of answers.

2. Scriptlets make it unnecessary for an agent to know that it "is in" a particular script in order to understand language inputs or real-world situations. There are two reasons why "being in" a script is a problematic starting point. First, automatically determining which script one is in can be extremely difficult since life doesn't proceed according to neatly packaged scripts. Second, people are often dealing with multiple topics at the same time, meaning they are, in effect, "in" different scripts simultaneously. For example, you can be cooking dinner while talking about buying a new lawn mower, grabbing the phone, and training the dog to not jump on the counter. Since scriptlets are not tied to a domain, the agent can use them as applicable across domains.

3. Manual knowledge engineering has gotten a bad rap because human knowledge is so extensive that the work of recording it can appear overwhelming. And, indeed, that would be true if we were to make it impossible to record *any* ontological knowledge about something unless we were prepared to record *all* ontological knowledge about it. However, that is an impractical and unnecessary point of departure – and certainly not the way that people learn about the world.

4. In order for agents to successfully engage in lifelong learning, they need to be able to acquire knowledge piecemeal and keep that knowledge organized so that both they and human developers can interpret the knowledge bases. Scriptlets allow for this.

*4.3. Scriptlets are the building blocks of domain scripts and are reused across different domain scripts.* They have a similar role in ontology as constructions have in lexicon: they combine chunks of information that can be used and reused across contexts and situations – in the spirit of object-oriented programming.

*4.4. Ontological knowledge, including scriptlets, should be recorded at the highest possible level of abstraction, making it maximally applicable across domains.* To motivate this generalization, let us consider an excerpt (for reasons of space) from the COMMUNICATIVE-ACTs subtree of the LEIA's ontology, shown in Figure 4. Brackets indicate children and * means 'copy the name of the parent here'. Each concept in the ontology expands to a frame that includes dozens of properties (AGENT, THEME, BENEFICIARY, INSTRUMENT, LOCATION, TIME, etc.), and whose property-facet-value triples are inherited unless locally overridden.

```
COMMUNICATIVE-ACT
  REQUEST-INFO <REQUEST-INFO-YN , REQUEST-INFO-WH, REQUEST-INFO-CHOICE-QU>
  REQUEST-ACTION
  PROPOSE-PLAN
  DECLARATIVE-SPEECH-ACT
  PERFORMATIVE-SPEECH-ACT <APOLOGIZE, EXPRESS-DOUBT, DOWNPLAY-MISTAKE, PRAISE, …>
  EMOTIONAL-COMMUNICATIVE-ACT <EXPRESS-DISPLEASURE, EXPRESS-PLEASURE, EXPRESS-UNCERTAINTY>
  RESPOND-TO-REQUEST-INFO
    RESPOND-TO-YN-REQUEST-INFO < *-POSITIVELY, *-NEGATIVELY, *-DON'T-KNOW>
    RESPOND-TO-WH-REQUEST-INFO <*-CONFIDENTLY, *-UNCONFIDENTLY, *-DON'T-KNOW>
    RESPOND-TO-REQUEST-INFO-CHOICE <*-CONFIDENTLY, *-UNCONFIDENTLY, *-DON'T-KNOW>
  RESPOND-TO- REQUEST-ACTION < *-POSITIVELY, *-NEGATIVELY, *-DON'T-KNOW>
  RESPOND-TO-PROPOSE-PLAN <ACCEPT-PLAN, REJECT-PLAN, RESPOND-TO-PLAN-UNCERTAINLY>
  BACKCHANNEL-SIMPLE
  CHECK-UNDERSTANDING
  SEEK-CLARIFICATION
  EMPTY-CONTENT-SPEECH-ACT
  HEDGE-BUY-TIME
  REFUSE-TO-RESPOND
  TELL-A-JOKE
```

Figure 4. An excerpt from the COMMUNICATIVE-ACT subtree of the ontology.

At the COMMUNICATIVE-ACT level of the ontology, the key knowledge elements that must be locally specified – and then inherited by all the descendants of the COMMUNICATIVE-ACT concept – are as follows: (a) the *default* for what happens after a COMMUNICATIVE-ACT is a COMMUNICATIVE-ACT by the other person; and (b) it is perfectly normal (recorded under the *sem* facet) for the other person to either carry out some non-verbal action (e.g., laugh, grimace, set off to do what he was told to) or do nothing at all, just listen quietly.

```
COMMUNICATIVE-ACT
  AGENT         default   HUMAN-#1
  BENEFICIARY   default   HUMAN-#2
  HAPPENS-NEXT  default   COMMUNICATIVE-ACT-#1 (AGENT HUMAN-#2) (BENEFICIARY HUMAN-#1)
                sem    or EVENT-#1 (AGENT HUMAN-#2)
                          DO-NOTHING-#1 (AGENT HUMAN-#2)
```

Moving down the COMMUNICATIVE-ACT subtree, the REQUEST-INFO level further constrains expectations about what will happen next: by default, a request for information will be followed by a response to the request for information, but it could also be followed by some other event. It would be quite odd to do nothing at all in response to a REQUEST-INFO (in contrast to a person simply making a statement), so that option is not inherited.

```
REQUEST-INFO
  AGENT         default   HUMAN-#1
  BENEFICIARY   default   HUMAN-#2
  HAPPENS-NEXT  default   RESPOND-TO-REQUEST-INFO-#1  (AGENT HUMAN-#2) (BENEFICIARY HUMAN-#1)
                sem       EVENT-#1 (AGENT HUMAN-#2)
```

Finally, at the level of a *yes-no* question, there are even more constrained expectations about what will happen next, as was shown in the knowledge structure presented in point 4.2 above.

*4.5. The computational lexicon provides the link between ontological events, including communicative acts, and the language utterances that instantiate them.* In many approaches to language processing, the distinction between ontology and lexicon is blurry or absent and/or the utterances covered are so few that there is no generalized approach at all. However, the interaction between ontology and lexicon is central to the theory of Ontological Semantics underlying language processing in OntoAgent, as is broad coverage of language inputs.

Consider again the concept REQUEST-INFO-YN. It is instantiated by a large number of constructions in the lexicon that link syntactic realizations of such questions with their semantic analyses. For each syntactic realization – e.g., *Did Subj VP? Can Subj VP? Is NP the $N_1$ who$_1$ VP? Is it possible that Clause? Clause, right? Clause, didn't Subj?, etc.* – dynamic semantic interpretation requires leveraging both static knowledge and procedural routines, all of which are recorded in the associated lexical sense. The left column of Table 1 shows one such construction.

*Table 1.* Lexical senses for a *yes-no* construction and an elliptical 'yes' response.

| A lexical sense for one yes/no construction | A lexical sense for one use of elliptical 'yes' |
|---|---|
| (do-aux-47<br>  (def "phrasal 'Do/did Subj VP?'")<br>  (ex "Did you eat a cookie?")<br>  (syn-struc<br>   ((root $var0) (cat aux))<br>    (subj ((root $var1) (cat n)))<br>    (vp ((root $var2) (cat v)))<br>    (punct ((root $var3) (root *quest-mark*) (cat punct)))<br>  (sem-struc<br>   (refsem1<br>    (REQUEST-INFO-YN<br>     (AGENT (value *speaker*))<br>     (BENEFICIARY (value *hearer*))<br>     (THEME  (value refsem2.<u>value</u>)))<br>   (refsem2<br>    (modality<br>     (type epistemic)<br>     (scope (value refsem3)))<br>   (^$var3 (null-sem +)))<br>  (meaning-procedures<br>   ; *presented informally for readability's sake*<br>   (LOCAL<br>    1. compute ^refsem3 using ^$var1 and ^$var2<br>    2. insert ^refsem3 into instantiated sem-struc<br>    3. contextually ground speaker and hearer)<br>   (EXTERNAL<br>    1. refsem2.value is waiting to be filled in by next<br>      utterance<br>    2. sense preference for next input is the default<br>      filler of HAPPENS-NEXT of refsem1))) | (yes-adv-6<br>  (def "used as fragmentary response to yn quest.")<br>  (ex "Did you eat a cookie?" "Yes.")<br>  (syn-struc<br>   ((root $var0) (cat adv)<br>    (punct ((root $var1) (cat punct) (root *period*))))<br>  (sem-struc<br>   (RESPOND-TO- REQUEST-INFO-YN-POSITIVELY<br>    (AGENT (value *speaker*)<br>     (BENEFICIARY (value *hearer*)<br>    (THEME (value refsem1))<br>    (refsem1<br>     (modality<br>      (type epistemic)<br>      (scope (value refsem2))<br>      (value 1)))<br>  (meaning-procedures<br>   ; *presented informally for readability's sake*<br>   (LOCAL<br>    1. identify refsem2, the proposition from the<br>      immediately preceding context whose scoping<br>      modality value slot is empty (which was posted<br>      to attention)<br>    2. Contextually ground speaker and hearer: e.g.,<br>      "Did you$_1$ X?" "Yes" [I$_1$ did X]))) |

The *syn-struc* zone defines the construction syntactically and allows for any shape of subject and verb phrase – both of which can be arbitrarily complex. The *sem-struc* zone is headed by the speech act REQUEST-INFO-YN, whose AGENT is the speaker and BENEFICIARY is the hearer. The THEME of the REQUEST-INFO-YN is the heart of this construction: it is the as-yet unknown value of epistemic modality scoping over the proposition. Different responses ("yes", "no", "probably not", shaking one's head, shrugging one's shoulders, etc.) will yield different values of this modality.[9] The scope of that modality is the meaning of the proposition, which needs to be composed by dynamically combining the meaning of the subject and the meaning of the verb phrase. The last line in the sem-struc zone indicates that the meaning of the question mark has already been taken care of – it need not be analyzed further.

The *meaning-procedures* zone calls procedural semantic functions. In the actual lexical senses, these are written as function calls with human-readable comments; but here, for readability's sake, we simply describe what they do. This zone is divided into local and external segments.

Local meaning procedures are needed to dynamically compute the contextual meaning of the input covered by the sense. In the left-side column of Table 1, the first local function call composes the meaning of the basic proposition, extracted from the question, by combining the meanings of the subject and the verb phrase, no matter their shape or complexity. (As McShane & Nirenburg (2021) shows, this process can be very complex.). The second local function inserts that propositional meaning into the instantiation of the rest of the sem-struc. For our example this results in the meaning representation:

```
REQUEST-INFO-YN-1
    AGENT           HUMAN-1
    BENEFICIARY     HUMAN-2
    THEME           MODALITY-1.value
MODALITY-1
    type            epistemic
    scope           INGEST-1
INGEST-1
    AGENT           HUMAN-2
    THEME           COOKIE-1
    scope-of        MODALITY-1
    TIME            (< find-anchor-time)
```

The final local function contextually grounds the speaker and the hearer in agent memory.

External meaning procedures provide anticipatory information to the attention module (cf. point 4.6 below). The first external function posts the information that what is expected next is the filling of the as-yet empty value of epistemic modality. The second function tells the agent to (a) look up in the ontology the concept that heads the sem-struc (here, REQUEST-INFO-YN), (b) determine the default filler of its HAPPENS-NEXT slot (i.e., RESPONSE-TO-REQUEST-INFO-YN), and (c) prefer interpretations of the next input (which might be an utterance and/or a gesture) that are headed by a concept in the subtree rooted in RESPONSE-TO-REQUEST-INFO-YN. Returning to Table 1, column 2, "Yes." is one such response since it is headed by RESPOND-TO-REQUEST-INFO-YN-POSITIVELY, which is a child of RESPOND-TO-REQUEST-INFO-YN. Looking more closely at the "Yes." sense, we see that it, too, includes meaning procedures which take care of the non-trivial bookkeeping of effectively translating (for our example) "Did you eat a cookie?" into "I ate a cookie" such that all objects, along with the coreferential event, are correctly grounded in agent memory. For our example, this will result in the following response text meaning representation:

```
RESPOND-TO-REQUEST-INFO-YN-POSITIVELY-1
    AGENT           HUMAN-2
    BENEFICIARY     HUMAN-1
    THEME           MODALITY-1
MODALITY-1
    TYPE            EPISTEMIC
    SCOPE           INGEST-1
    VALUE           1
INGEST-1
    AGENT           HUMAN-2
    THEME           COOKIE-1
    TIME            (< find-anchor-time)
```

---

[9] This is not the place to open up the issue of how epistemic modality interacts with beliefs.

*4.6 Different roles in different applications require the agent to operate at different depths of knowledge and reasoning.* Consider, for example, the difference between an agent that is tasked to observe and learn from watching others (an *observing agent*) versus an agent that is tasked to actively participate in an event (a *participating agent*). Whereas the observing agent can operate with ontological scriplets and a streamlined control mechanism, the participating agent needs fully specified domain scripts and full plan- and goal-oriented reasoning. To explain these options, we must introduce a few features of the OntoAgent cognitive architecture.

OntoAgent is comprised of 5 modules: perception, perception interpretation, attention and reasoning, action specification, and rendering (see McShane & Nirenburg, 2021, Ch. 7, p. 287; Nirenburg, McShane & English, 2021). All signals between processing modules are in the form of ontologically-grounded meaning representations of the same basic format that accommodate different perception modalities. For example, just as text meaning representations record the results of language interpretation, vision meaning representations record the results of vision interpretation. The processing of signals is coordinated by the agent's attention module. Any new input signal from perception is first sent to attention, which dispatches it to the correct interpretation module. The interpreted knowledge then goes back to attention and on to various situational reasoning modules. When attention receives any meaning representation, it has two choices, roughly corresponding to System 1 (instinctive) and System 2 (deliberative) processing.

Returning to our agents, different implementations of anticipatory processing are appropriate for when they are observing vs. participating in dialogs. When the observing agent interprets inputs (typically, in order to learn about the world), the information resulting from anticipatory meaning procedures is posted to attention without invoking goal- and plan-oriented reasoning. This allows the agent to understand (a) whether the communication is progressing in an expected way, and (b) how to efficiently resolve elliptical utterances, resulting in their full interpretations. That is, given many different senses of 'yes' in the lexicon, it will know to zero in on one or the other (which will have different procedural routines associated with them) based on the speech act it is expecting.

The observing agent doesn't need to invoke plans or goals for the language processing per se because it isn't being called on to decide how to respond. By contrast, a participating agent needs not only full goal- and plan-oriented reasoning, it also needs full-fledged domain scripts to guide its decision-making about how to respond, since scriptlets only provide the scope of options, not how to choose among them or what content to supply when responding using a particular kind of speech act.

*4.7 The opticon supports vision processing in the same way as the lexicon supports language processing.* We have been emphasizing throughout that speech acts are functionally no different from any other actions in terms of knowledge engineering and agent modeling. Just as language inputs need to be interpreted into TMRs, other kinds of perceptual inputs need to be interpreted into their version of XMRs. Consider the example of vision inputs, since they are central to our main topic of dialog interactions. If a person responds to a question by nodding or shrugging, that visual input needs to be translated into a VMR that plays the exact same role in agent cognition as do TMRs. Vision interpretation invokes an opticon that is formally parallel to the lexicon in that it includes a "syntactic" zone, a semantic zone, and an optional meaning procedures zone. For example, the opticon must contain a sense of "nodding one's head" that is identical to the "yes" entry in Table 1 except for the "syntactic" zone – which must be whatever the vision experts decide is appropriate. (In configuring demonstration systems to date, we have hand-coded the work of the vision processing that would call given opticon senses.) Corresponding approaches to knowledge representation must be developed for other channels of perception as well, such as non-linguistic audition, haptics, etc.

*4.8 Knowledge representation solutions are general-purpose.* We have motivated the utility of scriptlets using examples from dialog modeling. However, we did not decide to implement scriptlets solely to support dialog processing. In fact, this decision has a much broader impact. To give just one example (of many), consider CHANGE-EVENTs, which are events whose meaning involves the increase or decrease of a property value – e.g., accelerating, getting hotter, or losing weight. Before we introduced the notion of scriptlets, recording these property-value comparisons in the ontology resulted in awkward formulations, so we recorded them in inordinately fine-grained lexical senses for each associated word (McShane, Nirenburg & Beale, 2008). Although this served the purpose, it was methodologically not optimal, particularly since these explanatory lexical senses would need to be copied into the lexicons for all languages. Having adopted the notion of scriptlet, we can save effort and increase inspectability by recording these comparisons in the ontology. That is, the CHANGE-EVENT subtree shows that some property value is being compared between the PRECONDITION and EFFECT of the event, and each child – e.g., ACCELERATE – locally specifies which property is in question (VELOCITY) and the direction of the comparison (the value in EFFECT is GREATER-THAN the value in PRECONDITION).

CHANGE-EVENT
   PRECONDITION   default   PROPERTY-#1 (DOMAIN OBJECT-#1)
   EFFECT         default   PROPERTY-#2 (DOMAIN OBJECT-#1) (COMPARISON-RELATION PROPERTY-#1)
ACCELERATE
   PRECONDITION   default   VELOCITY-#1 (DOMAIN PHYSICAL-OBJECT-#1)
   EFFECT         default   VELOCITY-#2 (DOMAIN PHYSICAL-OBJECT-#1) (GREATER-THAN VELOCITY-#1)

Now the lexical sense for words like *accelerate* need only indicate that *X accelerates* means that *X* is the AGENT (for animates) or THEME (for inanimates) of the event ACCELERATE; and what that actually means is available in a scriptlet in the ontology.

## 6. Wrapping Up

Our goal is to create intelligent agents that can operate in any domain, learn through their operation, explain their operation in human terms, and become increasingly sophisticated over time. We believe that the best hope of doing this involves methodological and strategic decision-making that avoids near-term simplifications that will not hold up over time. We believe that such conceptual, modeling and implementational generalizations will lead to the development of agents that will not be disposed of every time a new capability is required of them. This, to our minds, is a pressing goal in building the next generation of intelligent agents. The introduction of scriptlets contributes to making this goal more attainable by streamlining the necessary knowledge engineering tasks.

## Acknowledgements

This research was supported in part by Grant #N00014-19-1-2708 from the U.S. Office of Naval Research. Any opinions or findings expressed in this material are those of the authors and do not necessarily reflect the views of the Office of Naval Research.